\documentclass[letterpaper, 10 pt, conference]{ieeeconf}  

\IEEEoverridecommandlockouts                              

\usepackage{graphicx}
\usepackage{cite}
\usepackage{amsmath,amssymb,amsfonts}
\usepackage{algorithmic}
\usepackage{graphicx}
\usepackage{textcomp}
\usepackage{xcolor}
\usepackage[ruled,vlined,linesnumbered]{algorithm2e}
\usepackage{bm}
\usepackage{subfig}
\usepackage{url}
\newcommand\figsize{0.46}

\title{\LARGE \bf Real-Time Fast Marching Tree for Mobile Robot Motion Planning in Dynamic Environments*
\thanks{*This research was supported in part by the Natural Science and Engineering Research Council of Canada (NSERC) under grant number DNDPJ 533392-18, General Dynamics Land Systems (Canada), and Defence R\&D Canada (DRDC).}
}

\author{Jefferson~Silveira$^{1}$\thanks{$^{1}$ J.\ Silveira is with the Department of Electrical \& Computer Engineering and the Ingenuity Labs Research Institute, Queen's University, Kingston, ON K7L 3N6 Canada \texttt{jefferson.silveira@queensu.ca}}, 
        Kleber~Cabral$^{2}$\thanks{$^{2}$K.\ Cabral is with the School of Computing, Queen's University, Kingston, ON K7L 3N6 Canada \texttt{kleber.cabral@queensu.ca}}, 
        Sidney~Givigi$^{3}$\thanks{$^{3}$S.\ Givigi is with the School of Computing and the Ingenuity Labs Research Institute, Queen's University, Kingston, ON K7L 3N6 Canada \texttt{sidney.givigi@queensu.ca}}
        and~Joshua~A.~Marshall$^{4}$\thanks{$^{4}$J.\ Marshall is with the Department of Electrical \& Computer Engineering and the Ingenuity Labs Research Institute, Queen's University, Kingston, ON K7L 3N6 Canada \texttt{joshua.marshall@queensu.ca}}
        }
        
\begin{document}

\maketitle
\thispagestyle{empty}
\pagestyle{empty}

\begin{abstract}

This paper proposes the Real-Time Fast Marching Tree (RT-FMT), a real-time planning algorithm that features local and global path generation, multiple-query planning, and dynamic obstacle avoidance. During the search, RT-FMT quickly looks for the global solution and, in the meantime, generates local paths that can be used by the robot to start execution faster. In addition, our algorithm constantly rewires the tree to keep branches from forming inside the dynamic obstacles and to maintain the tree root near the robot, which allows the tree to be reused multiple times for different goals. Our algorithm is based on the planners Fast Marching Tree (FMT*) and Real-time Rapidly-Exploring Random Tree (RT-RRT*). We show via simulations that RT-FMT outperforms RT-RRT* in both execution cost and arrival time, in most cases. Moreover, we also demonstrate via simulation that it is worthwhile taking the local path before the global path is available in order to reduce arrival time, even though there is a small possibility of taking an inferior path. 

\end{abstract}


\section{INTRODUCTION}

Planning optimal paths for mobile robots in environments filled with obstacles is a complex task that may require considerable computation time.  This can cause issues in time-sensitive applications, such as mining \cite{tian2021trajectory} because the longer it takes to execute a task, the lower the efficiency and the higher the expenses. This is also true for search and rescue applications\cite{hayat2020multi} since any saved time is used to look for survivors. These are applications where it is possible for a robot to encounter dynamic obstacles such as humans or other robots. To deal with such scenarios, this paper proposes a new path planning algorithm that works in real-time, thus reducing unnecessary waiting time while still avoiding dynamic obstacles.


\begin{figure}
\centerline{
\subfloat[]{
\includegraphics[width=0.45\linewidth]{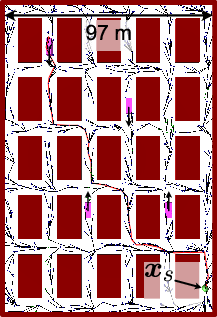}
\label{fig:mine_a}}\\
\subfloat[]{
\includegraphics[width=0.45\linewidth]{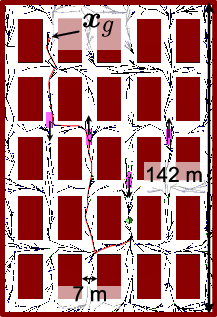}
\label{fig:mine_b}}
}
\caption{Illustration of a mine-like space. The start and goal configurations are shown as $x_s$ and $x_g$, respectively. The solution is highlighted in red. The pink squares are the dynamic obstacles representing mining trucks and the dark solid obstacles are the mine pillars. The goal is to traverse the space with the least amount of time to maximize profit.}
\label{fig:mine}
\end{figure}

In dynamic environments, new paths must be generated whenever a path becomes unfeasible. It is possible to save planning time by incorporating a re-planning approach that leverages the current search structure (e.g., a graph) that has been created previously instead of starting a new instance from scratch. One common approach to reducing planning time for time-sensitive applications 
is to generate a sub-optimal but complete path and start the path execution as soon as possible while it continues to optimize the path. Algorithms that fall into this category are, for example, the anytime planners \cite{karaman2011anytime, xu2020informed}. A different approach that can be applied when the application has hard time constraints is to expand the search until it is time to perform an action. In this case, actions are taken before the planner finds the complete trajectory. Algorithms that fall into this category are called \emph{real-time planners} \cite{naderi2015rt}.

While real-time planners provide the fastest trajectory execution, which may allow the robot to arrive at the destination sooner, there are two problems associated with these techniques. At the time to perform an action, the planner applies a heuristic function to decide the best local path to follow with the currently available information. Local minima in the heuristic function can cause the robot to visit unnecessary states. In addition, the planner may lead a robot to a state of unavoidable collision if there is not enough look-ahead \cite{cserna2016anytime}.

These problems can be greatly reduced if the planner is able to quickly search through the space to provide sufficient look-ahead for the real-time planner to select low-cost local paths while it searches for the optimal global path. Fast search is a characteristic of probabilistic sampling-based planners such as RRT* \cite{karaman2011sampling} and FMT* \cite{fmt}, which have been successfully applied in robotic problems for high-dimensional configuration spaces due to the speed at which these algorithms are able to search over the configuration space \cite{karaman2011sampling}. These types of planners are clearly good candidates for real-time applications. To fill this gap, the real-time variant RT-RRT* was proposed by Naderi et al. \cite{naderi2015rt}. This variant is capable of leveraging the generated tree to avoid dynamic obstacles by rewiring the tree around them as the robot executes the generated path. In addition, it generates local paths for execution before the complete path is found for real-time execution. 

Unfortunately, RRT variants have a common drawback. In environments with many obstacles, there is a low probability that a sample will be added to the tree due to collisions, requiring more iterations for the planner to find a solution. In contrast, the FMT* algorithm starts the search with all samples already distributed in the environment. As a result, FMT* provides solutions in less time and with lower costs than RRT*. This effect is discussed in more detail in \cite{fmt}. 

Inspired by the advantages of FMT* over RRT*, this paper proposes RT-FMT\footnote{We decided to drop the ``*'' in FMT* when naming our approach since real-time execution on local paths does not guarantee optimality}, a real-time variant of the FMT* that features local path generation, multiple-query planning, and dynamic obstacles avoidance. During the search, our approach quickly looks for the global solution and, in the meantime, generates local paths that can be used by the robot to start execution faster. In addition, our algorithm constantly updates the tree to avoid dynamic obstacles, and to maintain the tree root near the robot, which allows the tree to be reused multiple times for different goals. Fig.~\ref{fig:mine} shows the RT-FMT in a layout similar to an idealized underground mine.  Fig.~\ref{fig:mine_a} shows the initial stage with the solution highlighted in red, and Fig.~\ref{fig:mine_b} shows the environment some time later. We can see that the tree root is near the robot in both images and that the path has been updated to avoid the nearest truck (pink rectangle) in Fig.~\ref{fig:mine_b}. 

The performance of RT-FMT was evaluated against RT-RRT* in two dynamic environments. The results (see Section \ref{sec:results}) show that robots applying RT-FMT consistently arrived at the goal state before RT-RRT*. In addition, and they also show that when the robots start an early execution, risking moving towards sub-optimal paths, they consistently arrived at the goal state faster when compared to waiting for the complete optimal path.

The remainder of this paper is structured as follows. Section \ref{sec:related} presents some related works on planning in dynamic environments with online execution. Section \ref{sec:algorithm} presents the description of the proposed algorithm. Section \ref{sec:methodology} presents the evaluation methodology. Section \ref{sec:results} presents and discusses the results of the simulated experiments, and Section \ref{sec:conclusions} presents some final remarks and future work.

\section{RELATED WORK}
\label{sec:related}
In this section, we review the literature on online and real-time sampling-based planners for dynamic environments. Although there are algorithms that plan with prior information on the trajectories of the obstacles such as \cite{grothe2022st}, we focus on the techniques that only require the current position of the dynamic obstacles because it is easier to obtain such information from sensors or communication.

The CL-RRT \cite{luders2010bounds} is an approach that integrates an online controller with RRT in order to plan in the presence of uncertainty and still satisfy bounds on tracking error. When the robot moves along the generated path, the algorithm prunes infeasible branches. However, it does not update the root of the tree as RT-RRT* does. In addition, it has been shown in \cite{naderi2015rt}, that RT-RRT* is able to find shorter paths with fewer iterations in simpler systems without motion constraints. Online versions of the RRT* (ORRT*) and FMT* (OFMT*) have been proposed by Chandler and Goodrich \cite{chandler2017online}. Their approach is similar to RT-RRT* in the sense that multiple goals can be assigned without having to re-plan from scratch, but they apply a different rewiring approach. While RT-RRT* rewires around obstacles by checking for collisions, ORRT* and OFMT* apply a time-varying cost that increases near obstacles, causing nodes in the tree to prefer connections far from the obstacles. Another difference between these algorithms is the local path generation process that is part of RT-RRT*, which is not present in ORRT* and OFMT*. More recently, Tong et al.\ \cite{tong2019rrt} proposed RRT*FN-Replan, which is an online algorithm that maintains a fixed number of nodes in the tree and that takes advantage of previously generated tree branches to update the current plan. They have shown that their approach outperforms not only RT-RRT* but also ORRT*. 

Based on the current state of the art, we decided to implement RT-FMT and compare it against RT-RRT* for a few reasons. First, it has been shown to outperform CL-RRT. ORRT* and OFMT* do not include local path generation, and the obstacle avoidance module (based on cost) guarantees safe execution at the expense of higher cost trajectories. Even though it has been shown that RRT*FN-Replan outperforms RT-RRT*, these experiments involved testing only the rewire and re-plan processes, disregarding the first stage of tree expansion. Because the arrival time is directly affected as well by the tree expansion, it was deemed important to also consider this information.


\section{THE REAL-TIME FAST MARCHING TREE ALGORITHM}
\label{sec:algorithm}
In general, RT-FMT works by expanding a tree, similarly to how FMT* does, while it also checks for dynamic obstacles and rewires the tree around them. During the search, the algorithm also searches for the local paths with the least costs. These paths are used for the robot to start moving before planning is finished. When the robot reaches a new waypoint in the path, the root of the tree is updated. This event triggers a complete rewire of the tree to update the costs of all nodes. In a real-time application, the robot sends velocity commands at specific intervals. Therefore, we only allow the tree expansion and rewiring to run for a defined number of iterations ($N_e$) to expand and rewire the tree. This method can also be easily adapted to run for a desired time interval or planning frequency. 

The method is shown in Algorithm \ref{alg:main}, which starts by sampling $N$ configurations in free space (Line \ref{alg:samplefree}) considering only the fixed obstacles. This function also calculates the neighborhood radius $r_n$ based on the number of samples and the dimensionality of the problem according to
\begin{equation}
\label{eq:rn}
    r_n = \gamma_s 2\bigg(1 +\frac{1}{d}\bigg)^{\frac{1}{d}}\bigg(\frac{\mu(\mathcal{X}_{\rm free})}{\zeta_d} \bigg)^\frac{1}{d} \bigg(\frac{\log(N)}{N} \bigg)^\frac{1}{d},
\end{equation}
where $\gamma_s > 1$ is a tuning parameter, $d$ is the dimension of the problem, $\mu(\mathcal{X}_{\rm free})$ is the Lebesgue measure of the free space, and $\zeta_d$ is the volume of a unit ball in $\mathbb{R}^d$. Although FMT* provides an equation for $r_n$, we use the equation defined in \cite{karaman2011sampling} for PRM* since it computes an $r_n$ slightly bigger than the FMT* equation for $r_n$.

\begin{algorithm}[tb]
\SetKwFunction{SampleFree}{SampleFree}
\SetKwFunction{UpdateContext}{UpdateContext}
\SetKwFunction{ExpandFMT}{ExpandFMT}
\SetKwFunction{RewireFromObstacles}{RewireFromObstacles}
\SetKwFunction{RewireFromRoot}{RewireFromRoot}
\SetKwFunction{GeneratePath}{GeneratePath}
\SetKwFunction{UpdateRoot}{UpdateRoot}

 $\mathcal{T} \leftarrow \bm x_{s}$;  $\bm x_{\rm root} \leftarrow \bm x_{s}$\;
 $\mathcal{S} \leftarrow \SampleFree(N) \cup \bm x_{s} \cup \bm x_{g}$\; \label{alg:samplefree}
 $\mathcal{V}_{b} \leftarrow \emptyset$; $\mathcal{Q}_{o} \leftarrow \emptyset$;  $\mathcal{Q}_{r} \leftarrow \emptyset$\;
 $\mathcal{V}_{unv} \leftarrow \mathcal{S}\backslash\{\bm x_{s}\}$; $\mathcal{V}_{\rm open} \leftarrow \{\bm x_{s}\}$; $\mathcal{V}_{\rm closed} \leftarrow \emptyset$\;
 $\bm z \leftarrow \bm x_{s}$\;
 \While{True}{
 ($\bm x_{\rm robot} ,\bm x_{g}, \mathcal{N}_{b}) \leftarrow \UpdateContext(\mathcal{T}, \mathcal{X}_{\rm Dobs})$\; \label{alg:updateContext}
 \For{$i=1$ \KwTo $N_{e}$}{
  $\ExpandFMT(\mathcal{T})$\; \label{alg:expandAndRewire}
  $\RewireFromObstacles(\mathcal{T})$\; \label{alg:rewireObs}
  $\RewireFromRoot(\mathcal{T})$\; \label{alg:rewireRoot}
}
 $(\bm x_{\rm root}, \bm x_1, ..., \bm x_k) \leftarrow \GeneratePath(\mathcal{T}, \bm x_{\rm root})$\; \label{alg:generatepath}
 \If{$\bm x_{\rm robot}$ \upshape{is near} $\bm x_{\rm root}$}{
    $\bm x_{\rm root}\leftarrow \bm x_{1}$\;
    $\UpdateRoot(\mathcal{T},\bm x_{\rm root})$\; \label{alg:updateroot}
 }
 Steer robot towards  $\bm x_{\rm root}$\; \label{alg:steer}
 Perform other tasks\;
}
\caption{RT-FMT($\bm x_{s}, \bm x_{g}, \mathcal{X}_{Fobs}, \mathcal{X}_{\rm Dobs}, N_s, N_e $)}
\label{alg:main}
\end{algorithm}

Inside the infinite loop, the algorithm updates the context of the problem by returning the current position of the robot and goal $x_{g}$, and by finding the nodes in the tree that have been blocked or unblocked by the dynamic obstacles. When a configuration $\bm x_q$ in the tree is blocked, its cost $\textup{c}(\bm x_q)$ is set to infinity. When a node is unblocked, its cost is updated to
\begin{equation}
    \textup{c}(\bm x_q) = \textup{c}(\bm x_{\rm parent}) + \textup{Cost}(\bm x_{\rm parent}, \bm x_{q}), \label{eq:cost}
\end{equation}
where $\bm x_{\rm parent}$ is the configuration of the parent of $\bm x_q$, and
\begin{equation}
    \textup{Cost}(\bm u, \bm v) = ||\bm v- \bm u||.
\end{equation}
If the updated node has children, its children's costs are recursively updated. Lines \ref{alg:expandAndRewire}\textendash\ref{alg:rewireRoot} expand the tree according to Algorithm \ref{alg:ExpandFMT}, and rewire the tree based on Algorithm \ref{alg:RewireFromObstacle}. Line \ref{alg:generatepath} returns a global path if $\bm x_g$ is in the tree or a local path otherwise. The local path is found by computing 
\begin{equation}
    \bm x_k \leftarrow \arg\min_{\bm x \in \mathcal{T}}\textup{c}(\bm x) +  ||\bm x -\bm x_{goal}||,
\end{equation}
and then a path starting at $x_{\rm root}$ and ending at $\bm x_k$ is generated. While we do not limit $k$, RT-RRT* generates a path up to a specific $k$. More details on the local path generation can be found in Algorithm 6 in Naderi et al.~\cite{naderi2015rt}. The approach resembles the A* search. 

In Line \ref{alg:updateroot}, the root of the tree is set to the next configuration in the path, which is always the second element since the path always starts at the old root.  Finally, the algorithm steers the robot towards the new root of the tree on Line \ref{alg:steer}. If there are other tasks to perform such as mapping, and localization, they can be called in the main function as well.

\subsection{Expanding the Tree}

The tree expansion is described  in Algorithm \ref{alg:ExpandFMT}. Most of the algorithm (Lines \ref{alg:x_near}\textendash\ref{alg:findz}) was inspired by FMT*, proposed in \cite{fmt}, but our implementation has two major differences. 

First, the loops in the original implementation were substituted for conditional statements. These statements ensure that only one node can be added in the tree per call. If multiple nodes are added to the tree at once, there is a possibility of delaying other tasks since the processor will spend too much time expanding many nodes at once. In addition, Line \ref{alg:checkdynamic} not only checks for fixed obstacles but also checks whether $\bm y_{\rm min}$ is not being blocked by a dynamic obstacle according to (\ref{eq:cost}). 

Second, in the original approach, once a node has checked all possible connections with its neighbors, it is closed and never checked again. In our approach, we do not spend time expanding the nodes that are inside $\mathcal{X}_{\rm Dobs}$. As a consequence, when a dynamic obstacle moves, there will be unvisited nodes around closed nodes. To add these nodes to the tree, our algorithm must be able to reopen closed nodes nearby. This is done by adding all $\bm z$ that are near unvisited nodes at closing time to $\mathcal{V}_{\rm toOpen}$ (Line \ref{alg:ztoopen}). Then, when the regular expansion is finished (Line \ref{alg:znull}), the algorithm reopens these nodes (Line \ref{alg:reopen}) to continue the expansion in case a dynamic obstacle moves.

\begin{algorithm}[tb]
\SetKwFunction{Near}{Near}
\SetKwFunction{PopLast}{PopLast}
\SetKwFunction{Cost}{Cost}
\SetKwFunction{c}{c}
\SetKwFunction{CollisionFree}{CollisionFree}
\SetKwFunction{Open}{Open}
\SetKwFunction{Close}{Close}
\lIf{$\mathcal{X}_{near}$ = $\emptyset$ $\&$ $\bm z \neq \emptyset$}{$\mathcal{X}_{near} \leftarrow$ \Near($\bm z$,$\mathcal{V}_{unv}$)} \label{alg:x_near}
\Else{
$\bm x = \PopLast(\mathcal{X}_{near})$\; 
$\mathcal{Y}_{near} \leftarrow$ \Near($\bm x$,$\mathcal{V}_{\rm open}$)\;
\bm $y_{\rm min} \leftarrow \arg\min_{\bm y \in \mathcal{Y}_{near} }(\c(\bm y) + \Cost(\bm x, \bm y))$\;

\If{$\CollisionFree(\bm y_{\rm min}, \bm x) \& \c(\bm y_{\rm min}) < \infty$}{ \label{alg:checkdynamic}
$\mathcal{V}_{open, new} \leftarrow  \mathcal{V}_{open, new} \cup \{\bm x\}$\;
$\mathcal{V}_{unv} \leftarrow \mathcal{V}_{unv}\backslash\{\bm x\}$\;
$\c(\bm x) \leftarrow \c(\bm y_{\rm min}) + \Cost(\bm y_{\rm min}, \bm x)$\;
$\mathcal{T} \leftarrow \mathcal{T} \cup \{\bm x, (\bm y_{\rm min}, \bm x) \}$
}
\If{$\mathcal{X}_{near}$ = $\emptyset$ $\&$ $\bm z \neq \emptyset$}{
\Close($\bm z$)\;
$\mathcal{Z}_{near} \leftarrow$ \Near($\bm z$,$\mathcal{V}_{unv}$)\;
$\bm z \leftarrow \arg\min_{\bm y \in \mathcal{V}_{\rm open} }\c(\bm y)$\; \label{alg:findz}
\ForEach{$\bm x \in \mathcal{Z}_{near}$ }{ \label{alg:ztoopen}
\lIf{\CollisionFree($\bm z, \bm x$)}{$\mathcal{V}_{\rm toOpen} \leftarrow \mathcal{V}_{\rm toOpen} \cup \bm z  $} 
}

}
\If{$\bm z = \emptyset$}{ \label{alg:znull}
$\Open(\mathcal{V}_{\rm toOpen})$;~$\mathcal{V}_{\rm toOpen} = \emptyset$\; \label{alg:reopen}
$\bm z \leftarrow \arg\min_{\bm y \in \mathcal{V}_{\rm open} }\c(\bm y)$\; }
}

\SetKwProg{Def}{def}{:}{}
\Def{\Close($\mathcal{V}$)}{
$\mathcal{V}_{\rm open} \leftarrow  \mathcal{V}_{\rm open} \cup \mathcal{V}_{open, new}\backslash\mathcal{V}$\;
$\mathcal{V}_{\rm closed} \leftarrow  \mathcal{V}_{\rm closed} \cup \mathcal{V}$\;
}

\SetKwProg{Def}{def}{:}{}
\Def{\Open($\mathcal{V}$)}{
$\mathcal{V}_{\rm open} \leftarrow  \mathcal{V}_{\rm open} \cup  \mathcal{V}$\;
$\mathcal{V}_{\rm closed} \leftarrow  \mathcal{V}_{\rm closed} \backslash \mathcal{V}$\;
}
\caption{ExpandFMT($\mathcal{T}$)}
\label{alg:ExpandFMT}
\end{algorithm}

\subsection{Rewiring the Tree}

As the dynamic obstacles move around the environment, the tree nodes are constantly being blocked and unblocked by Line \ref{alg:updateContext} in Algorithm \ref{alg:main}. When a node is blocked or unblocked, its cost is changed and all its children are recursively updated. The task of the {\tt RewireFromObstacle} method in Algorithm \ref{alg:RewireFromObstacle} is to find the connections with lower cost in the neighborhood of nodes that have recently been blocked or unblocked. This function only rewires the nodes that have recently been affected by a dynamic obstacle. The rewiring process starts by adding all blocked nodes to $\mathcal{Q}_o$. Then, nodes are iteratively removed from the list and the algorithm tries to find parents nearby with a lower cost. If there is a connection with a lower cost that is also collision-free, the children of the updated nodes are also added to $\mathcal{Q}_o$. 

\begin{algorithm}[tb]
\SetKwFunction{PopFirst}{PopFirst}
\SetKwFunction{UpdateParentChild}{UpdateParentChild}
\SetKwFunction{RecalculateChildrenCost}{RecalculateChildrenCost}
\lIf{$\mathcal{Q}_o$ = $\emptyset$}{$\mathcal{Q}_o \leftarrow \mathcal{N}_{b}$} \label{alg:addblockedtolist}
\Else{
$\bm x_b \leftarrow \PopFirst(\mathcal{Q}_o)$\;
   \If{$\bm x_b \notin \mathcal{X}_{\rm Dobs}$}{
        $\mathcal{Y}_{near} \leftarrow \Near(\bm x_b,\mathcal{V}_{\rm open}\cup \mathcal{V}_{\rm closed}$)\;
        \bm $y_{\rm min} \leftarrow \arg\min_{\bm y \in \mathcal{Y}_{near} }(\c(\bm y) + \Cost(\bm x_b, \bm y))$\;
        \If{$\CollisionFree(\bm y_{\rm min}, \bm x) \& \c(\bm y_{\rm min}) < \infty$}{
        $\UpdateParentChild(\mathcal{T}, \bm y_{\rm min}, \bm x_b)$\;
        $\RecalculateChildrenCost(\bm x_b)$
        }
   }

}
\caption{RewireFromObstacles($\mathcal{T}$)}
\label{alg:RewireFromObstacle}
\end{algorithm}

As the robot moves around the environment, the tree root is updated by Algorithm \ref{alg:main}, Line \ref{alg:updateroot}. When this happens, the algorithm triggers the {\tt RewireFromRoot} function (Line \ref{alg:rewireRoot}) to rewire all nodes in the tree by inserting the new root into $\mathcal{Q}_{r}$. The {\tt RewireFromRoot} function is very similar to Algorithm \ref{alg:RewireFromObstacle}. The algorithm removes a node from $\mathcal{Q}_{r}$ and tries to find better connections in the tree. When a new connection is made, the children of the updated node are also added to $\mathcal{Q}_{r}$. This causes a chain reaction that starts from the root and updates all nodes in the tree. This method has also been implemented without any loops to only update a single node per call. Our implementation triggers this chain reaction whenever the root is updated. However, this can easily be modified to happen at a desired frequency.


\section{EVALUATION METHODOLOGY}
\label{sec:methodology}
\begin{figure}
\centerline{
\subfloat[RT-FMT]{
\includegraphics[width=0.4\linewidth]{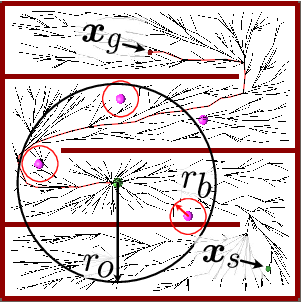}
\label{fig:maze_a}}\\
\subfloat[RT-RRT*]{
\includegraphics[width=0.4\linewidth]{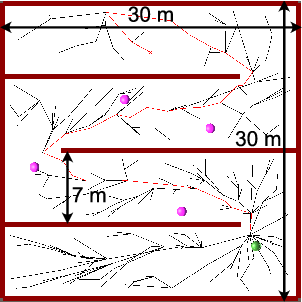}
\label{fig:maze_b}}
}
\caption{Simulation environment with the Maze space for (a) RT-FMT and (b) RT-RRT*. The parameter $r_o$ represents the sensing range of the robot, and the distance $r_b$ represents the safety radius in which tree nodes are considered blocked if a dynamic obstacle is within $r_o$ from the robot.}
\label{fig:maze}
\end{figure}

The main goal of our evaluation process was to study the advantages of real-time planning and execution using FMT over RRT in certain environments, as well as to verify whether real-time execution and planning can result in faster arrival times. To properly achieve our evaluation goals, we implemented and tested our algorithm in Unity 3D using a similar environment to the one provided by \cite{naderi2015rt}. Then, we prepared two different scenarios. A Maze space, shown in Fig.~\ref{fig:maze}, which was designed as a toy problem to highlight one shortcoming of RRT-based algorithms in maze-like spaces. This shortcoming is caused by the lower probability of successfully connecting a node to the tree without hitting a wall. The second environment is a Mine-like space illustrated in Fig.~\ref{fig:mine}, as an application problem that was designed following the room and pillar method of creating underground mines.

The experiments in the Maze space were executed with the robot traveling at 2 m$/$s, with $r_b = 2$ m, $r_o = 10$ m, and a robot radius of 0.5 m. For the Mine space, the robot traveled at 4 m$/$s, with $r_b = 14$ m, $r_o = 50$ m, and a robot radius of 1.5 m. For both environments, the dynamic obstacles traveled at half the robot's speed. All remaining parameters of RT-RRT* were kept the same as provided by \cite{naderi2015rt}. F or example, the number of tentative expansions per action ($N_e$) was set to 32. For RT-FMT, we also maintained  $N_e = 32$, and $\gamma_s$, defined in \eqref{eq:rn}, was kept at 1.1 for all experiments.

We split the evaluation methodology into three experiments with each one repeated 50 times for a different number of sample attempts. They ranged from 500 to 4500 samples with an increment of 1000 samples. It is important to emphasize that sample attempts is not the number of nodes in the tree. For RT-FMT, it indicates the free space sample count, and for RT-RRT*, the sample attempts represent the number of times the algorithm sampled a point and attempted to expand the tree. Since both approaches are very different in building the tree, care must be taken when analyzing the results. This same approach was applied by \cite{fmt} to compare FMT* against RRT*.

Experiment 1 involved obtaining the planning times, executed costs, and arrival times by running the planners in a non real-time manner; i.e., allowing the robot to move only after trying the defined number of samples and finding the complete path. These metrics were used as a baseline to judge the performance of the real-time execution for Experiments 2 and 3. Experiment 2 allowed the robots to execute the local paths in real-time without the presence of dynamic obstacles. With this experiment, it is possible to evaluate whether the robots can arrive earlier at the goal, considering that there is a risk of finding a costlier trajectory when compared to Experiment 1. Experiment 3 is similar to 2, but with dynamic obstacles present in the space. In the Maze space, the dynamic obstacles maintained a fixed initial position and they moved in random directions. In the Mine space, their vertical positions were randomly selected at the start of the simulation, and they would move down if they appeared in the top part of the environment, and vice-versa.

The RT-FMT implementation used for the evaluation procedure is available at \url{github.com/offroad-robotics/rt-fmt-icra.git}.

\section{SIMULATED EXPERIMENTS}
\label{sec:results}

This section presents the results obtained from the simulations of the different experiments and discusses the results. During the experiments, both algorithms reached quickly 100~\% success rate on both spaces during experiments 1 and 2 when increasing the number of sample attempts. During Experiment 3, which is the more challenging one, only RT-FMT reached 100~\% at 4500 samples in the Maze space, while RT-RRT* achieved only 46~\% at the same number of samples. For the Mine space, RT-FMT achieved 84~\% and RT-RRT only 50~\% success rate. The low success rate of RT-RRT* is caused by node connections with longer edges than RT-FMT, as shown in Fig. \ref{fig:maze}. 

During the execution of both algorithms, the robot does not react to changes in the environment when it is moving in between nodes. While it is possible to limit the edge lengths of RT-RRT* to improve the success rate of the algorithm, it worsens other important characteristics such as planning time and executed cost that are analyzed in this section. For such reasons, we did not modify this behavior in RT-RRT*.


\subsection{Experiment 1}

\begin{figure}
\centering
\subfloat[Maze space]{
\label{fig:experiment_1_a}
\includegraphics[width=\figsize\linewidth]{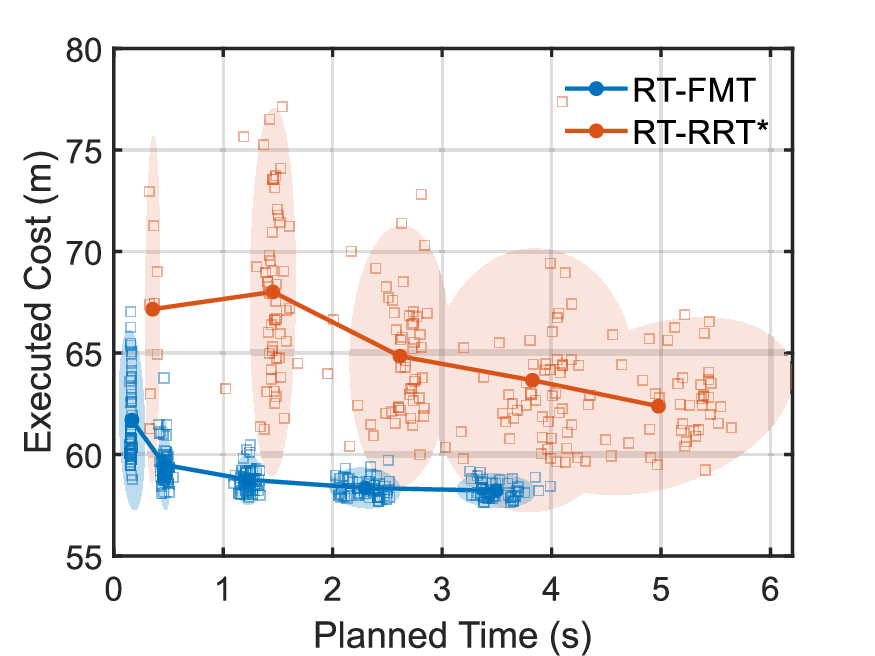}
}
\subfloat[Mine space]{
\label{fig:experiment_b}
\includegraphics[width=\figsize\linewidth]{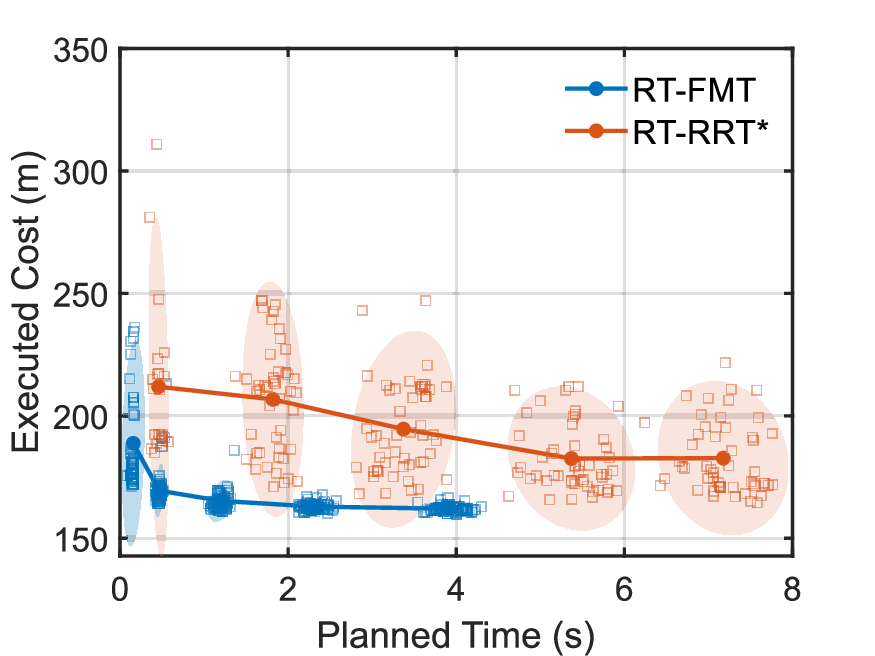}
}
\caption{Simulation results for Experiment 1. Hollow squares represent a single simulation, solid circles the average on both axes for different sample counts, and the ellipses 90~\% confidence level.}
\label{fig:experiment_1}
\end{figure}

For this experiment, we show the solution cost for all runs in Fig.~\ref{fig:experiment_1}. It is clear that RT-FMT outperforms RT-RRT* in planning time and solution cost for both environments. The solution was found faster with the blue curve located to the left side of the graph (shorter planning time) and below the orange line (shorter executed path). Also, the solution found by the RT-FMT yielded a shorter path for the same sample count. This is because, for the same number of sample count (i.e., attempts to connect points to the tree), RT-FMT will have many more nodes added to the tree than RT-RRT*, which is a direct consequence of how the samples of RRT* and FMT* are computed.

Furthermore, the planning time is much more predictable in RT-FMT than RT-RRT* due to its smaller deviation, which indicates that RT-FMT is better suited for real-time applications due to the hard real-time bounds that are present in these systems~\cite{cserna2016anytime}.

\subsection{Experiment 2}

\begin{figure}
\centering
\subfloat[]{
\label{fig:maze_2_a}
\includegraphics[width=\figsize\linewidth]{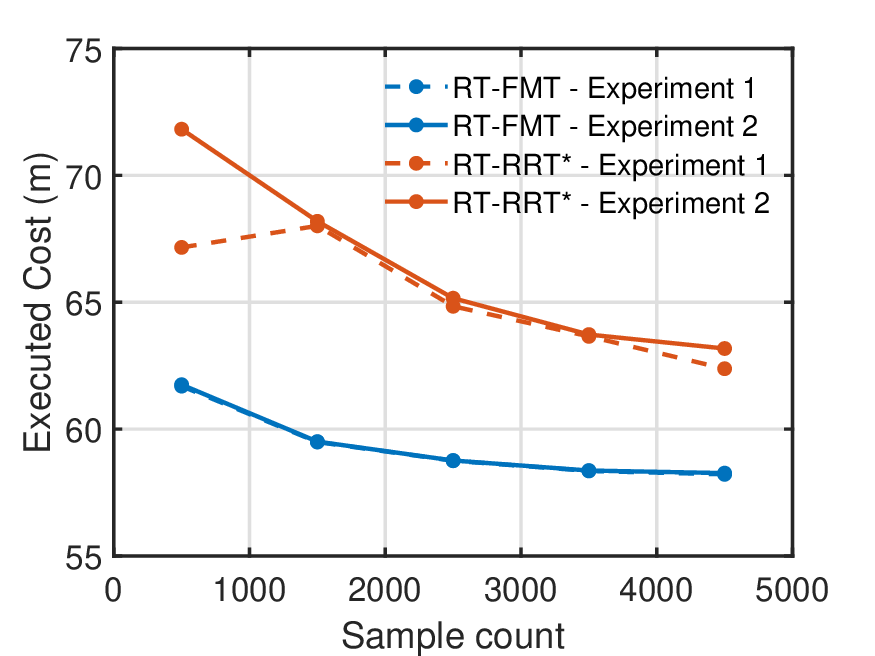}
}
\subfloat[]{
\label{fig:maze_2_b}
\includegraphics[width=\figsize\linewidth]{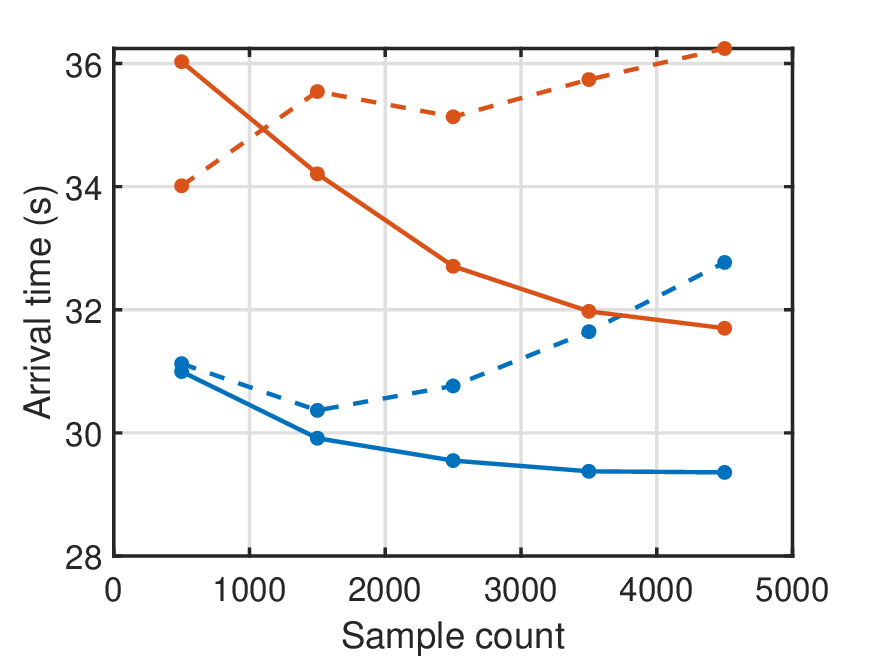}
}
\caption{Simulation results for the Maze in Experiment 2.}
\label{fig:maze_2}
\centering
\subfloat[]{
\label{fig:mine_2_a}
\includegraphics[width=\figsize\linewidth]{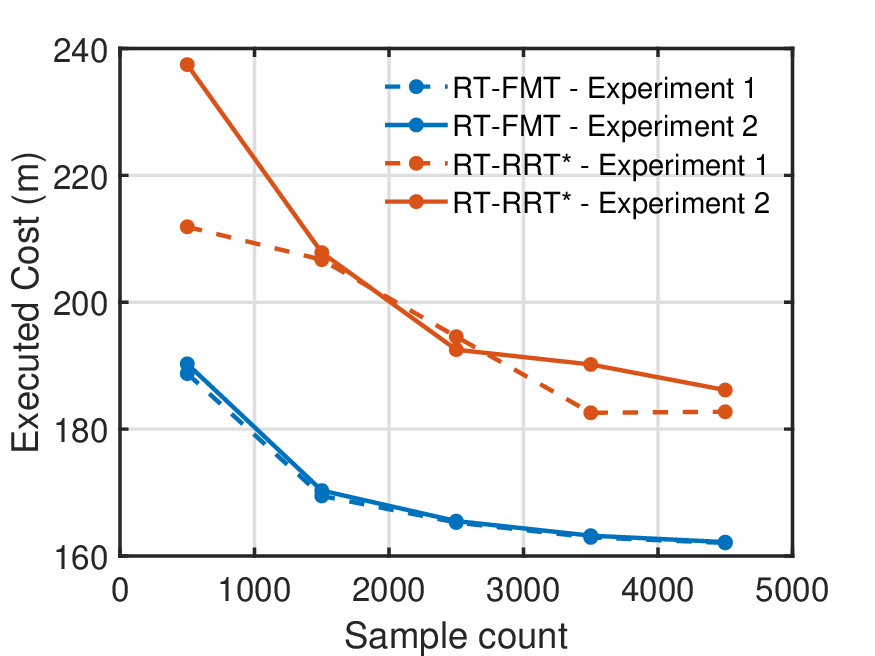}
}
\subfloat[]{
\label{fig:mine_2_b}
\includegraphics[width=\figsize\linewidth]{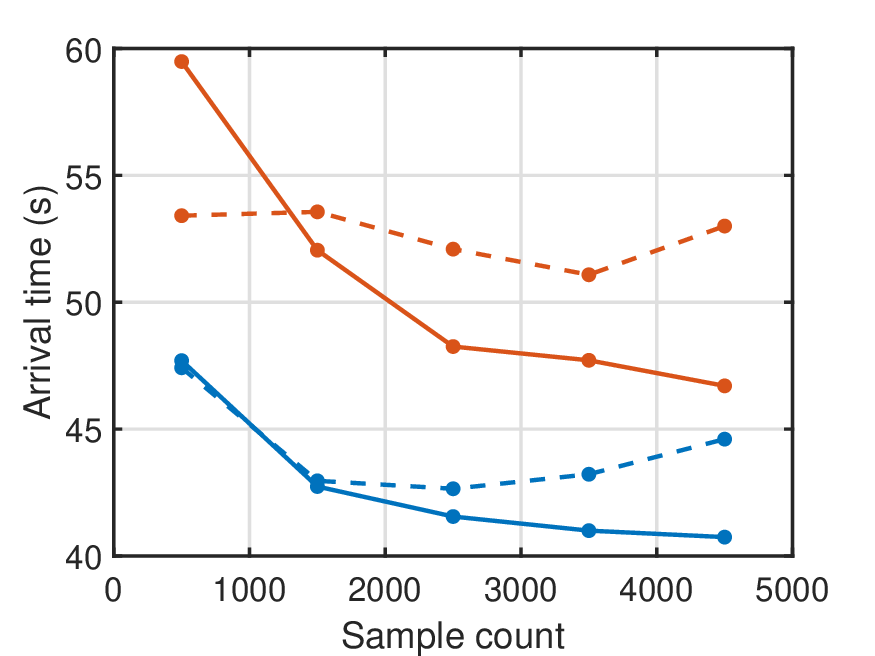}
}
\caption{Simulation results for the Mine in Experiment 2.}
\label{fig:mine_2}
\end{figure}

For the results of this experiment and Experiment 3, instead of displaying the planning time as in Fig.~\ref{fig:experiment_1}, we display the sample count in the $x$-axis because it is fixed for each simulation group. The results of this experiment are shown in Fig.~\ref{fig:maze_2} and Fig.~\ref{fig:mine_2} for the Maze and Mine spaces, respectively. From these images, we can draw a few relevant conclusions. First, the executed cost and arrival time of RT-FMT was also smaller than RT-RRT* for both environments. And, interestingly, RT-FMT was able to match the ideal cost (from Experiment 1) almost perfectly, as can be seen in Fig.~\ref{fig:maze_2_a} and Fig.~\ref{fig:mine_2_a}. RT-RRT* also performed relatively well.

Another relevant observation to mention in Fig.~\ref{fig:maze_2_b} and Fig.~\ref{fig:mine_2_b} is the fact that the arrival time for Experiment 1 starts to monotonically increase after 1500 samples. This is caused by the extra time the algorithm takes to expand the tree with more samples. As a consequence of this fact and of both algorithms being able to closely match the executed costs to Experiment 1, both approaches had smaller arrival times for higher sample counts (dashed lines in  Fig.~\ref{fig:maze_2_b} and Fig.~\ref{fig:mine_2_b}). This result indicates that it is indeed worthwhile risking taking a sub-optimal path, in the beginning, to save planning time since the difference in executed cost was small for RT-RRT* and negligible for RT-FMT.

\subsection{Experiment 3}

\begin{figure}
\centering
\subfloat[]{
\label{fig:maze_3_a}
\includegraphics[width=\figsize\linewidth]{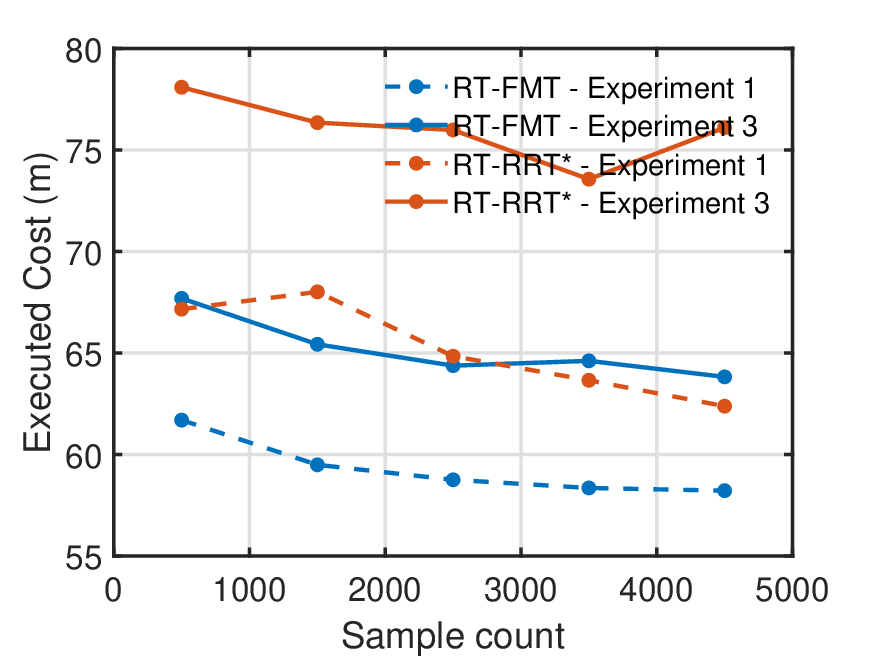}
}
\subfloat[]{
\label{fig:maze_3_b}
\includegraphics[width=\figsize\linewidth]{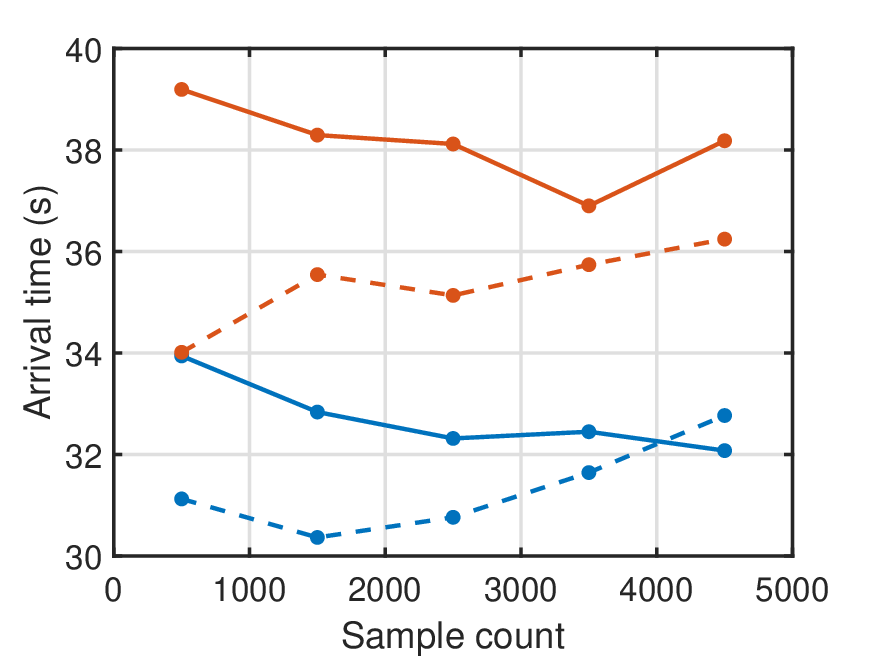}
}
\caption{Simulation results for the Maze in Experiment 3.}
\label{fig:maze_3}
\centering
\subfloat[]{
\label{fig:mine_3_a}
\includegraphics[width=\figsize\linewidth]{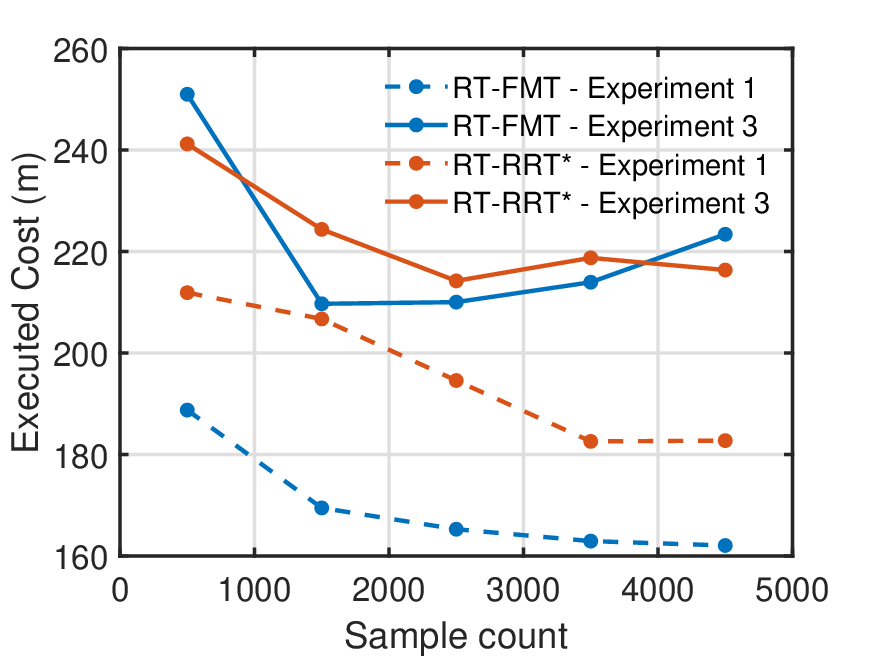}
}
\subfloat[]{
\label{fig:mine_3_b}
\includegraphics[width=\figsize\linewidth]{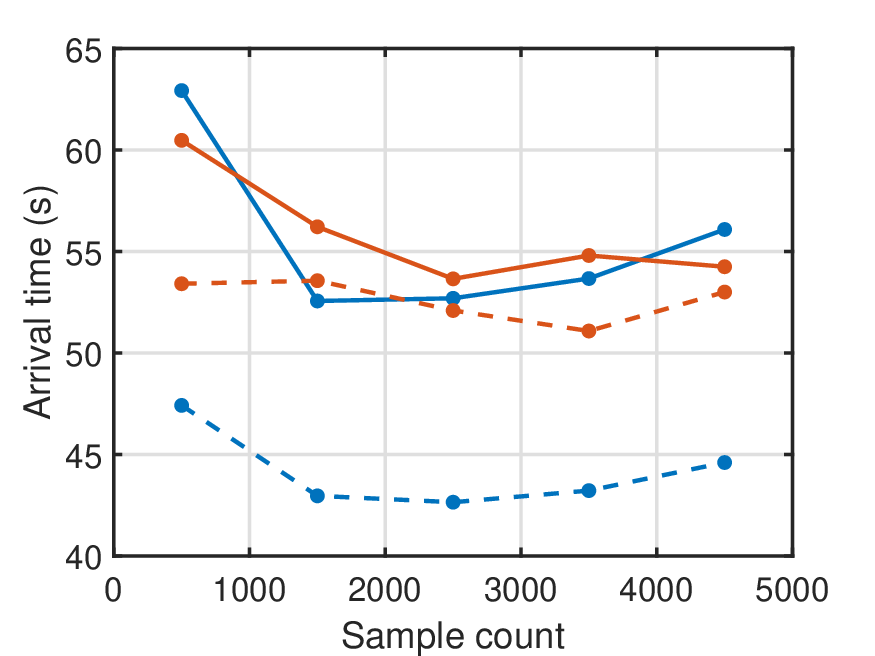}
}
\caption{Simulation results for the Mine in Experiment 3.}
\label{fig:mine_3}
\end{figure}

After adding the dynamic obstacles, the cost of the executed path, unsurprisingly, increased with respect to the execution without obstacles (Experiment 2). This is shown in both  Fig.~\ref{fig:maze_3} and Fig.~\ref{fig:mine_3}. However, that was the only common characteristic between the results of the Maze and Mine spaces for Experiment 3. 

If we focus on the arrival time of the maze space for RT-FMT (Fig.~\ref{fig:maze_3_b}), we see that as the number of samples increased, the difference in arrival time of Experiment 3 and Experiment 1 became smaller up to the point that the real-time execution arrived faster at the goal state, even with the extra task of avoiding dynamic obstacles. This is an excellent result since it is very common to select a very high number of nodes to guarantee that a solution will be found.

Regarding the Mine space, the big difference in executed cost of Experiment 3 and Experiment 1 in Fig.~\ref{fig:mine_3} can be explained by looking at the layout of the mine in Fig.~\ref{fig:mine}. When a dynamic obstacle blocks the robot in the middle of a hallway, the robot has to go around the block to avoid it, which greatly increases the executed cost. That is also the reason why RT-FMT and RT-RRT* performed similarly, they both had to go around the same number of blocks, on average, to avoid the obstacles.

Another behavior that is important to discuss is the upward trend of the arrival cost as the number of samples increased in Fig.~\ref{fig:mine_3_b} for RT-FMT. This was a consequence of the Mine space layout, the sample count definition in Section \ref{sec:methodology}, and how RT-FMT and RT-RRT* sample points in the environment. The Mine space contains more occupied space than free space, which caused the RT-FMT samples to be densely packed since they are guaranteed to lie in the small free space. In contrast, most sampled points for RT-RRT* failed to connect to the tree because of the likelihood of collisions. As a result, RT-FMT had many more nodes in the tree than RT-RRT*. With many nodes densely packed, the rewiring process of RT-FMT became too slow to respond to the dynamic obstacles in advance, causing the robot to react to dynamic obstacles as if it was short-sighted, which, consequently, also forced the robot to take longer routes. The same effect would have happened to RT-RRT* if the tree had a similar number of nodes to RT-FMT. Still, RT-FMT had an 84~\% success rate while RT-RRT* had only 50~\%.

\section{CONCLUSIONS}
\label{sec:conclusions}

This paper proposed RT-FMT, which is a real-time path planning algorithm that is capable of quickly generating low-cost paths in environments with dynamic obstacle avoidance with unknown trajectories. The real-time capability of the algorithm comes from the fact that it generates local paths for the robot to follow while the global path is not available. All these characteristics indicate that RT-FMT is ideal for time-critical applications. To show the capabilities of RT-FMT, we compare it against RT-RRT* on simulated environments with dynamic obstacles. The results show that RT-FMT has higher success rates, lower traveled distances, and smaller arrival times in almost all cases. The greater performance of RT-FMT over RT-RRT* is mostly related to the fact that, in cluttered spaces, the probability of randomly connecting a node to the tree is very low for RT-RRT*, which requires many attempts to grow the tree. On the other hand, RT-FMT starts the search with all nodes already laid out in the environment, which reduces the number of attempts to expand the tree.


Future work on the proposed algorithm will involve an implementation in the Open Motion Planning Library (OMPL) \cite{sucan2012the-open-motion-planning-library} to facilitate comparisons with other state-of-the-art approaches, mainly on higher dimensional problems, and to understand the effects of the real-time variation on the complexity of the algorithm and its computing time. The OMPL implementation might also simplify tests on real robots. In addition, a variation of the approach that works for constrained robots can also be implemented.



\addtolength{\textheight}{-15cm}   



\end{document}